# Non-Ideal Program-Time Conservation in Charge Trap Flash for Deep Learning

Shalini Shrivastava[a], Vivek Saraswat[a,*], Gayatri Dash[a], Samyak Chakrabarty[b], Udayan Ganguly[a, c]

[a]Department of Electrical Engineering, Indian Institute of Technology Bombay, Mumbai - 400076, Maharashtra, India

[b]Department of Electronics and Electrical Communications Engineering, Indian Institute of Technology Kharagpur, Kharagpur – 721302, West Bengal, India

[c]The Centre for Semiconductor Technologies (SemiX), IIT Bombay, Mumbai – 400076, Maharashtra, India

*Corresponding Author: svivek@ee.iitb.ac.in

**Abstract – Training deep neural networks (DNNs) is computationally intensive but arrays of non-volatile memories like Charge Trap Flash (CTF) can accelerate DNN operations using in-memory computing. Specifically, the Resistive Processing Unit (RPU) architecture uses the voltage-threshold program by stochastic encoded pulse trains and analog memory features to accelerate vector-vector outer product and weight update for the gradient descent algorithms. Although CTF, offering high precision, has been regarded as an excellent choice for implementing RPU, the accumulation of charge due to the applied stochastic pulse trains is ultimately of critical significance in determining the final weight update. In this paper, we report the non-ideal program-time conservation in CTF through pulsing input measurements. We experimentally measure the effect of pulse width and pulse gap, keeping the total ON-time of the input pulse train constant, and report three non-idealities: (1) Cumulative $V_T$ shift reduces when total ON-time is fragmented into a larger number of shorter pulses, (2) Cumulative $V_T$ shift drops abruptly for pulse widths < 2 $\mu$s, (3) Cumulative $V_T$ shift depends on the gap between consecutive pulses and the $V_T$ shift reduction gets recovered for smaller gaps. We present an explanation based on a transient tunneling field enhancement due to blocking oxide trap-charge dynamics to explain these non-idealities. Identifying and modeling the responsible mechanisms and predicting their system-level effects during learning is critical. This non-ideal accumulation is expected to affect algorithms and architectures relying on devices for implementing mathematically equivalent functions for in-memory computing-based acceleration.**

Keywords – artificial neural networks, backpropagation, charge trap flash, non-volatile memories, resistive processing unit, stochastic gradient descent, trapping de-trapping

## 1. Introduction

Deep Neural Networks have enabled the solution of many complex cognitive tasks such as speech and object detection as well as pattern recognition [1]. The large availability of training data has allowed robust supervised learning algorithms like gradient descent by backpropagation to achieve excellent "near-human" accuracies for various neural networks [2,3]. However, these are data-intensive algorithms requiring large matrix product operations followed by network parameters update. To reduce data movement, in-memory computing techniques have been proposed for hardware acceleration of DNN operations [4]. During backpropagation [5], the network undergoes a forward pass which primarily requires vector-matrix inner product calculation ($z_i^l = \left(\Sigma_j W_{ij}^l . x_j^{l-1}\right)$) to generate output activations ($x_i^l = a\left(z_i^l\right)$, where $a$ is the activation function) for all layers ($l$). The final layer output is then compared to a target and the errors ($\delta_j$) are propagated in a backward pass. The forward activations and backpropagated errors are then used to calculate the weight gradients ($\Delta W_{ij}^{l} \propto -x_i^{l-1}.\delta_j^{l}$) and the weights are updated ($W_{ij} = W_{ij} + \Delta W_{ij}$). This operation typically requires vector-vector outer product calculation followed by reading out weights, adding the updates writing back the new weights to the memory array. While inner-product acceleration typically depends on Ohm's law and Kirchhoff's Current law to perform the multiply and accumulate operations, the in-memory parallelization of outer-product and weight update is more involved. To accelerate the latter, the Resistive Processing Unit (RPU) architecture was proposed specifically for the analog non-volatile accumulative memories [6,7]. The components of RPU for a CTF memory array (Fig. 1. (a)) implementing a layer of weights of a neural network are described as follows:

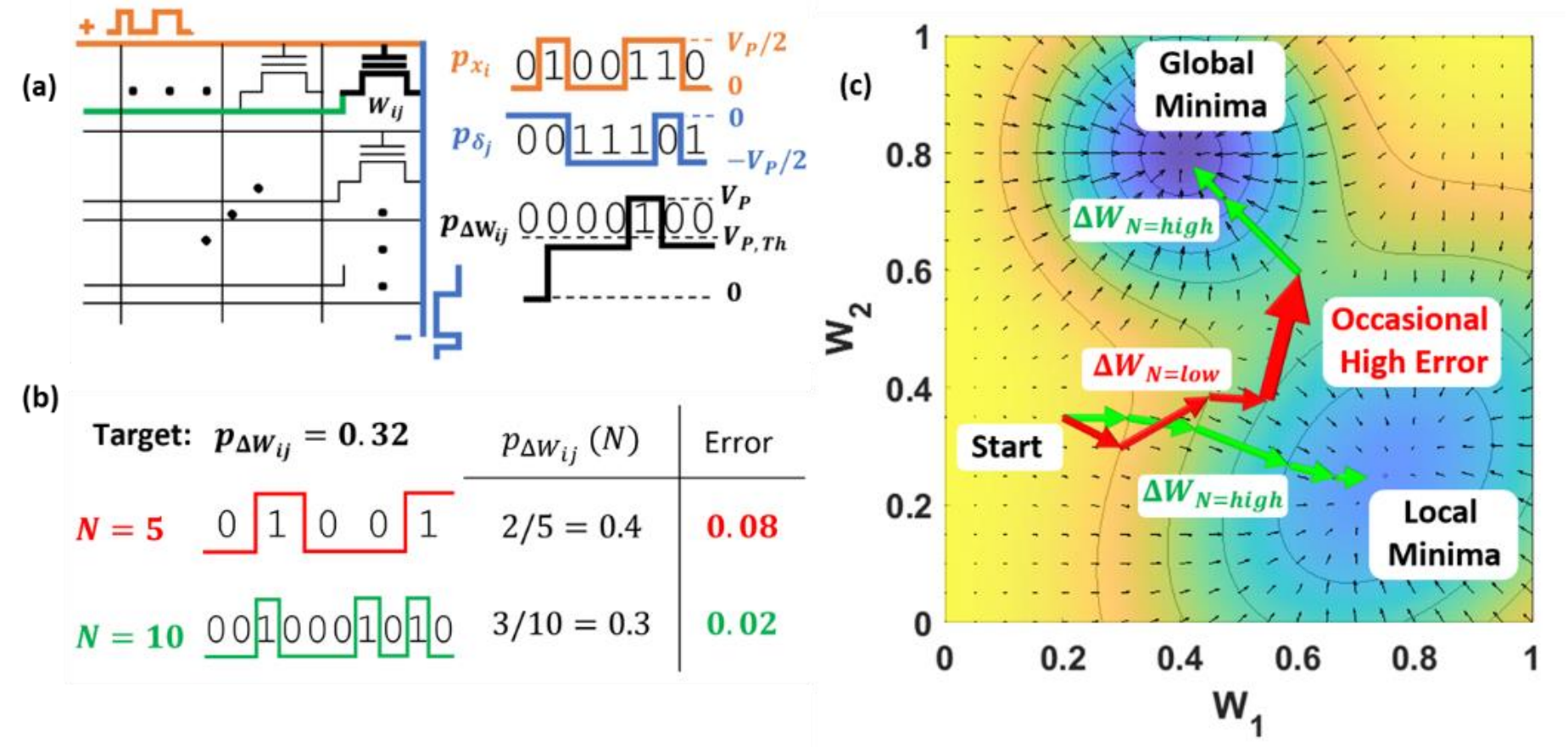

**Fig. 1. Signaling in RPU architecture:** (a) A CTF crossbar array with activations and errors encoded as rates of stochastic switching pulse trains, the CTF device responds to overlapped pulses due to programming threshold, (b) An example of how higher $N$ results in lower error variance in gradient calculation, (c) Training error as weights evolve, lower $N$ results in noisier gradient descent with potential for improved search space exploration.

1. Stochastic Pulse Encoding: The input activation ($x_i$) and error ($\delta_j$) are encoded as rates of uncorrelated and probabilistically switching time pulses ($p_{x_i}, p_{\delta_j}$) (Fig. 1. (a)). The pulses are applied to the rows and columns respectively where they interact with each weight located at cross points of the crossbar memory array. The total length of the ON/OFF pulse trains ($N =$ total number of ON and OFF slots) determines the precision with which the inputs and outputs can be represented (Fig. 1. (b)).
2. Multiplication and Update: Due to the nature of independent stochastic pulse trains, the overlapped row and column pulses effectively represent the multiplication of inputs and hence an encoding of the required weight update ($p_{\Delta W_{ij}}$) (Fig. 1. (a)). Most non-volatile memories have a non-linear writing mechanism with a voltage threshold. Thus, the memory cell responds only to overlaps for appropriately designed pulses. Further, the non-volatile memories can accumulate conductance change with each consecutive overlapped pulse, leading to the required weight update.

RPU allows $n^2$ multiplications and weight updates (for an $n \times n$ weight array) to all be parallelized and completed in one shot. The added complexity and source of error is the generation and use of stochastic pulse trains for representing weight update inputs. However, the controllability of error in gradient calculation by modifying the length of the stochastic pulse train allows for search space exploration vis-à-vis the accuracy in gradient descent (Fig. 1. (c)). For the memory technology implementing this architecture, the following are desirable [7]: (a) Linear programming with pulse number, (b) Gradual weight update ($\sim 1000$ levels), (c) High non-linearity with programming threshold and (d) Large conductance range ($G_{max}/G_{min}$) compared to the variations/noise in updates.

Emerging non-volatile memories such as Resistive RAM (RRAM) have shown different levels of linearity in weight updates [8]. However, the number of discrete conductance levels is less than 120. Phase Change Memory (PCM) requires a differential resistance of 2 PCM cells as only Set process is gradual while the Reset is abrupt, resulting in a larger footprint for 1000 linear levels of learning [9]. Ferroelectric RAM (FeRAM) has shown linearity only for 16 levels of learning [10]. CMOS-based RPU is proposed in [11] but occupies a large area with an added difference of memory volatility. CTF has also been essential in accelerating deep neural networks through in-memory computing techniques [6,12–14]. It offers a selector-less bit-cell owing to the three-terminal design with $V_T$ of the CTF transistor storing the weight (conductance level) and the source-drain current allowing the readout. Recently, SANOS based memory has been proposed for highly linear and gradual weight update; up to 1000 levels have been demonstrated [6,15]. The symmetry and analog nature are much superior to contemporary demonstrations from RRAM [8], PCM [9], and FeRAMs [10]. The $V_T$ change data due to pulsing has been modelled and used to predict system level classification accuracy using RPU based learning [6].

However, the accumulation of charge in CTF due to the applied pulse trains of different lengths ($N$) is ultimately significant in determining the final weight update. In this letter, we experimentally measure the effect of pulse widths and pulse gaps of the input pulse train keeping the total ON-time constant on the charge accumulation behavior of a SANOS CTF MOSCAP device [15]. We identify three aspects that require experimental investigation:

1. Can pulse width ($t_{pw}$) be reduced arbitrarily to get proportionately smaller weight updates – enable arbitrarily gradual weight update? A typical assumption in [6] is that weight update should depend upon total pulse-ON time, $T_{ON}$ (i.e., invariant with the number of divisions $N$). In short, is the charge accumulation conserved as $N$ changes if the total input time of programming remains same?
2. Longer pulse trains (higher $N$) of stochastic pulses reduce error variance in representing weight update ($\sigma_N \propto \frac{1}{\sqrt{N}}$, Fig. 1. (b)) [16]. For efficient gradient descent, $\sigma_N$ maybe tuned from smaller $N$ (more stochastic but fewer programming pulses) to larger $N$ (more precise but more programming pulses). One can engineer an efficient stochastic gradient descent with the schedule of weight updates with different levels of $N$-dependent "noisiness" like simulated annealing in optimization problems [17].
3. As the write pulses are a random stream of ON/OFF pulses with variable inter-pulse spacing (i.e., $t_{gap}$), we need to know whether the $V_T$ shift has any $t_{gap}$ dependence.

It has previously been assumed implicitly that $t_{pw}$ can be made arbitrarily smaller or $N$ and $t_{gap}$ dependence on $V_T$ shift is absent [6]. However, to accurately predict the system-level performance of algorithms running CTF models, it is essential to characterize these behaviors experimentally. First, we report that the charge accumulation in the CTF MOSCAP is not conserved when a long program pulse is divided into shorter pulses (higher $N$, smaller $t_{pw}$). For even shorter pulses, the $V_T$ shift vanishes altogether abruptly. However, reducing the $t_{gap}$ results in recovery of the reduced $V_T$ shift. Second, we use measurements on different process splits for the gate stack to present a physical mechanism based on traps in blocking oxide to explain the non-ideal charge accumulation behavior. It is critical to model the non-ideal $V_T$ accumulation effects and its impact on system-level neural network training performance, latency, and energy requirements. This system-level modelling is outside the scope of presented work, it will be published elsewhere.

# 2. Experimental Section

## 2.1. Device Fabrication

The schematic of the CTF device used for measurements is shown in Fig. 2. (a). The Silicon Nitride-based CTF has an n-Si substrate and a Boron-doped p+ source/drain regions. The gate stack consists of an $SiO_2$ **tunnel oxide (TO)** formed by Rapid Thermal Oxidation, an $Si_3N_4$ **charge trap layer (CTL)** formed by low-pressure chemical vapor deposition, an $Al_2O_3$ **blocking oxide (BO)** formed by chemical vapor deposition and an n+ poly-Silicon gate. The details of device fabrication are discussed in detail in [18].

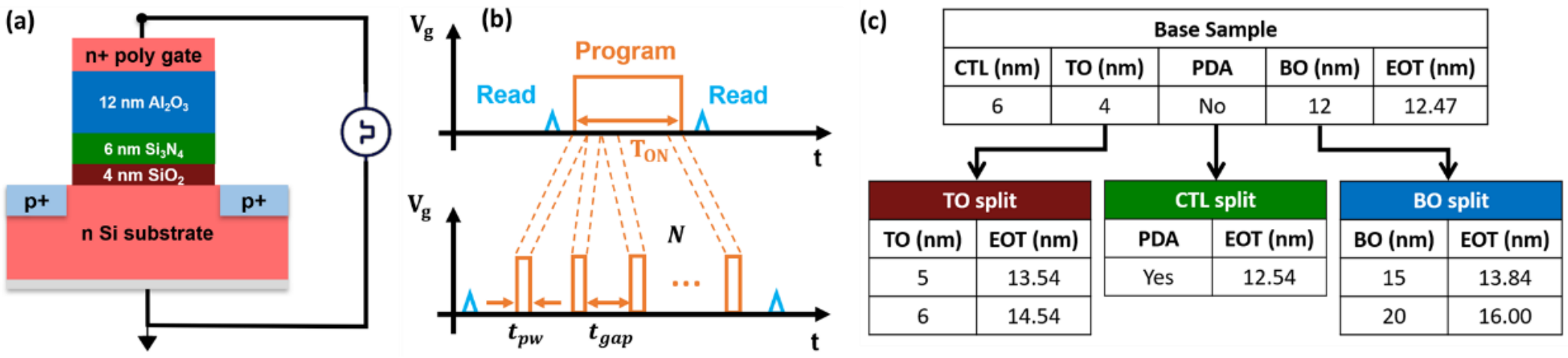

**Fig. 2. Device schematic and experiments:** (a) Device schematic, (b) Total ON-time, $T_{ON}$, is divided into $N$ pulses of $t_{pw}$ width and $t_{gap}$ spacing; pre- and post-programming $V_T$ are read, (c) Process splits for tunnel oxide (TO) layer thickness, post-deposition annealing (PDA) of charge trap layer, and the blocking oxide (BO) layer thickness.

## 2.2. Setup for $t_{pw}$ and $t_{gap}$ variation experiments

All voltage pulses are applied across the gate and substrate of the CTF device. The device is initialized using pre-programming erase pulses till a constant threshold voltage, $V_{T,0}$ is achieved before applying any program pulses. The initialized $V_{T,0}$ is read using a $[-2, 2]\ V$ sweep and $10\ KHz$ capacitance-voltage measurement using the constant capacitance, $C_{FB} = 15\ pF$, method. Next, a $T_{ON} = 2.5\ ms$ program pulse of magnitude 12.5 $V$ is applied and the $V_T$ is measured again. The total ON time, $T_{ON}$, can be divided into $N$ discrete pulses of the same magnitude and pulse width, $t_{pw}$, such that $T_{ON} = N \times t_{pw}$ (rise and fall times are excluded and fixed at 150 $ns$ for all pulses). Each consecutive pulse is separated by an OFF period from the next by $t_{gap}$ duration (Fig. 2. (b)). This experiment of initialization and programming is performed for different total number of divisions in the pulse train, i.e., $N$ as shown in the Table I.

Table I. Values of $N$ and $t_{pw}$ in experiments

| $\boldsymbol{N}$ | 1 | 10 | 25 | 50 | 100 | 500 | 1000 | 2000 | 5000 | 10000 |
|---|---|---|---|---|---|---|---|---|---|---|
| $\boldsymbol{t_{pw}(s)}$ | 2.5 m | 0.25 m | 0.1 m | 50 $\mu$ | 25 $\mu$ | 5 $\mu$ | 2.5 $\mu$ | 1.25 $\mu$ | 0.5 $\mu$ | 0.25 $\mu$ |

For each $N$, the $t_{gap}$ between consecutive pulses is kept fixed and chosen from 100 $ns$ to 10 $s$. The initial $V_{T,0}$ and final $V_{T,N}$ before and after the application of each pulse train is measured.

## 2.3. Process splits for gate stack

To understand the role of different layers in the gate stack on the $V_T$ shift during pulse train timing experiments; different process splits are created with respect to the base sample (Fig. 2. (c)). Three different layer thicknesses are chosen for tunnel oxide (TO) (4 - base, 5 and 6 nm). Similarly, three different layer thicknesses are chosen for blocking oxide (BO) (12 - base, 15 and 20 nm). There is also a split where the nitride charge trap layer (CTL) is subject to post-deposition annealing (PDA) in $O_2$ environment at 900°C, which is expected to change the stoichiometry to form some oxynitride. Different layers have different effective oxide thicknesses (EOT). It is ensured that all splits are programmed to the same $V_T$ on the application of a single pulse ($N = 1$, $t_{pw} = T_{ON} = 2.5$ ms) by appropriate biasing. Different $N$ pulse train experiment is repeated on all splits, and the initial and final $V_T$ shift before and after the application of each pulse train is measured.

## 3. Results and Discussion

### 3.1. $V_T$ shift in response to $t_{pw}$ and $t_{gap}$ variation

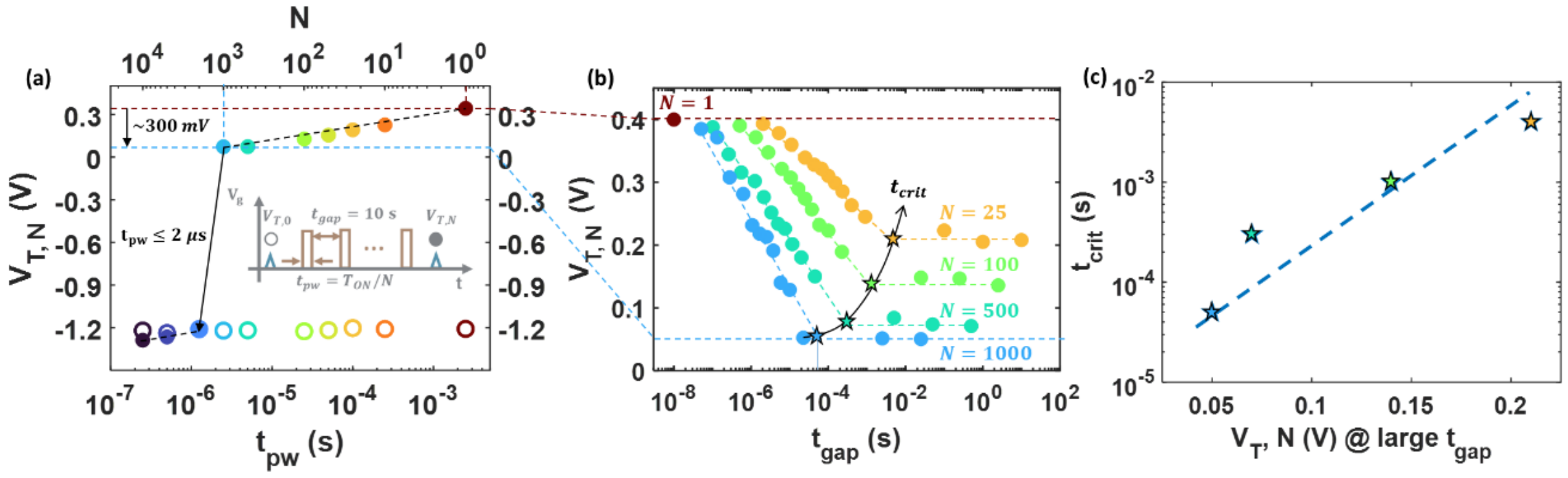

**Fig. 3. $V_T$ shift in response to $t_{pw}$ and $t_{gap}$:** (a) Initial $V_{T,0}$ (open markers) and final programmed $V_{T,N}$ (filled markers) for pulse trains of different $N$ (shown in the inset), a high $t_{gap} = 10\ s$ is used and $t_{pw} = T_{ON}/N$, where $T_{ON} = 2.5\ ms$. $V_{T,0}$ is constant for all pulse trains while $V_{T,N}$ reduces as $N$ increases and shows an abrupt fall for $t_{pw} \leq 2\ \mu s$, (b) Final $V_{T,N}$ for different $t_{gap}$ and $N$ saturates at high $t_{gap}$ and gets recovered as $t_{gap}$ reduces, (c) Estimated $t_{crit}$ as a function of the saturated $V_{T,N}$ at large $t_{gap}$.

The results of experiments described in Section 2.2, i.e., variation of $t_{pw}$ and $t_{gap}$ is shown in Fig. 3. For all input pulse trains, the starting $V_{T,0} \sim -1.2\ V$ is ensured to be constant. A high inter-pulse spacing ($t_{gap} = 10\ s$) is used first, and the final programmed $V_{T,N}$ is plotted for different $N$, maintaining $t_{pw} = T_{ON}/N$. Two main observations are listed below (Fig. 3. (a)):

1. A logarithmic fall is observed in $V_{T,N}$ ($\sim 300\ mV$) with $N$ as $N$ increases from 1 to 1000.
2. An abrupt fall is observed in $V_{T,N}$ for $t_{pw} \leq 2\ \mu s$ indicating that programming is disabled below a knee point.

Next, the $t_{gap}$ is reduced from $10\ s$ to $100\ ns$ and the following is observed (Fig. 3. (b)):

3. There is a strong logarithmic $t_{gap}$ dependence of $V_{T,N}$ when $t_{gap}$ is small. For $t_{gap}$ longer than a critical time scale $t_{crit}$, the $V_{T,N}$ saturates (Fig. 3. (b)). The $t_{crit}$ is higher for longer $t_{pw}$ (higher $N$) (Fig. 3. (c)).

We term these behaviors as the non-idealities in CTF accumulation. The telltale disablement of programming for $t_{pw} \leq 2 \mu s$ indicates a dead-zone period in programming for every pulse (Fig. 3. (a)). The charge injection into CTL should be significantly lower during this dead-zone and only increase after this period is over. Further, the charge injection is extremely sensitive to the electric field in the TO (exponential dependence) in the FN-tunneling regime. Thus, we hypothesize that if the TO electric field was low at the beginning of each pulse but increased to saturate at the end of the dead-zone per pulse, then the tunneling current (exponential with TO field) would amplify this electric field dependence, effectively producing a very non-linear and approximately step-like tunneling current. For identifying the source of dead-zone and tunneling field enhancement to validate this hypothesis, the $V_T$ shift behavior is next observed for different process splits.

### 3.2. $V_T$ shift in response to process splits for gate stack

#### 3.2.1. Sensitivity to BO splits as opposed to TO, and CTL splits

To generate a physics-based qualitative model for reduction in $V_{T,N}$, pulsing measurements are performed for different process splits as described in Section 2.3. The final programmed $V_{T,N}$ for different process splits of tunnel oxide (TO) thickness, post-deposition annealing (PDA) of charge trap layer (CTL) and blocking oxide (BO) thickness with increasing $N$ is shown in Fig. 4. (a)-(c). For all process splits, it is ensured that the same $V_{T,N}$ is achieved for $N = 1$ by appropriately choosing the bias magnitude. Fig. 4. (a) and (b) reveal the relative insensitivity of the reduction in $V_{T,N}$ with $N$ for TO and CTL splits. The $V_{T,N}$ reduction behaviors with $N$ are similar for different TO thicknesses and for CTL PDA cases compared to the base sample. However, Fig. 4. (c) demonstrates that BO splits significantly change the $V_{T,N}$ reduction behavior with $N$. Higher BO thickness splits witness higher $V_{T,N}$ reduction with $N$. The results of total $\Delta V_T = V_{T,N} - V_{T,0}$, before and after application of $N = 1000$ pulses for different splits are collated in Fig. 4. (d) as a function of effective oxide thicknesses (EOT) of the gate stack. The only splits which result in significant sensitivity of $\Delta V_T$ with EOT is the BO thickness splits. This indicates that the physical reason for reduced $V_T$ shift at higher $N$ or lower $t_{pw}$ is primarily associated with the nature of BO.

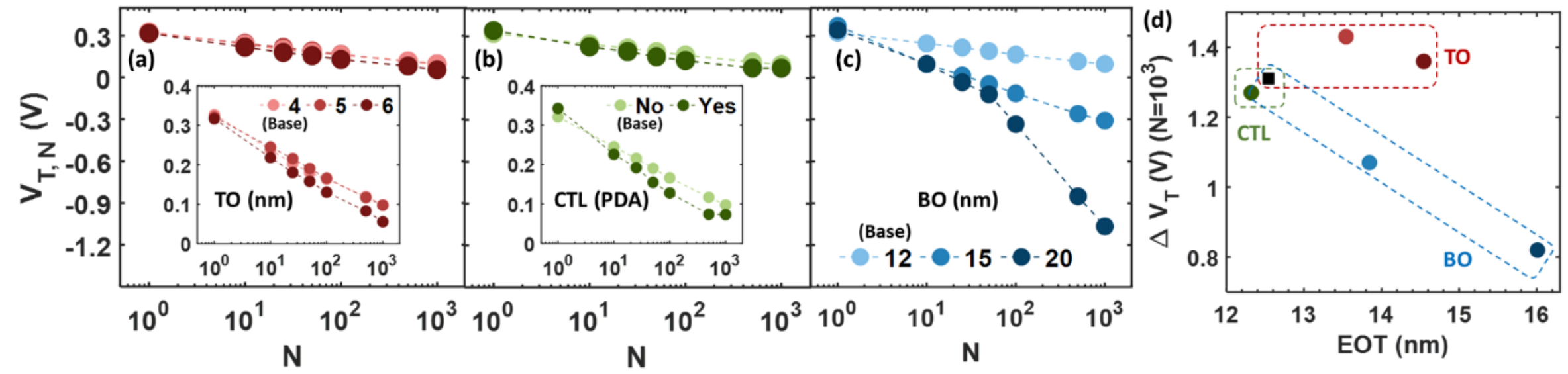

**Fig. 4. Sensitivity to blocking oxide splits:** Final programmed $V_{T,N}$ as $N$ increases for different process splits of (a) tunnel oxide (TO) thickness (inset – zoomed), (b) post-deposition annealing (PDA) of charge trap layer (CTL) (inset – zoomed), and (c) blocking oxide (BO) thickness (inset – zoomed), (d) $\Delta V_T = V_{T,N} - V_{T,0}$ for $N = 1000$ pulses as a function of effective oxide thickness for different process splits (black square is the base sample).

### 3.2.2. Qualitative model for the $V_T$ shift with $N$ and $t_{gap}$

Now, a qualitative model to account for the dead-zone and tunneling field enhancement (hypothesized at the end of Section 3.1.) can be built based on the trapping and de-trapping of positive charges in BO to explain the $V_T$ shift behaviors with $N$ and $t_{gap}$:

1. The traps in $Al_2O_3$ as BO are shallower and spread throughout the thickness of BO [19] and hence faster than the $Si_3N_4$ traps as CTL (which require tunneling past the TO). Hence, electrons can be ejected from the BO traps to the poly-Silicon gate at high program biases to leave behind a more positively charged BO. Fig. 5. (a) shows the qualitative band diagram through the gate stack when a programming pulse is applied. The dotted line shows the band diagram immediately after the field is applied. The electrons from the BO are able to transiently get ejected from the BO traps or equivalently positive charges get trapped at BO, which leads to a modified band diagram. These positive charges in the BO are volatile and are removed once the programming pulse is removed (unlike the charge stored in CTL).
2. When the programming pulse is applied, the positive charges in the BO further enhance the electric field in the TO (Fig. 5. (b)). This enhances the charge injection ($J_{tunnel}$) and hence charge storage in the CTL. The positive BO charge also prevents further storage in BO traps leading to saturation in the BO trapped charge.
3. The trapping of positive charges in BO has a finite timescale. This timescale creates an approximate dead-zone (shaded red in Fig. 5. (b)) at the start of every programming pulse when the TO field is relatively lower.
4. After the dead-zone period, the electron tunneling and trapping in CTL becomes significant (step-like behavior in $J_{tunnel}$ before and after the dead-zone period). This stored charge in CTL is responsible for the $V_T$ changes measured after the entire pulse train has been applied.

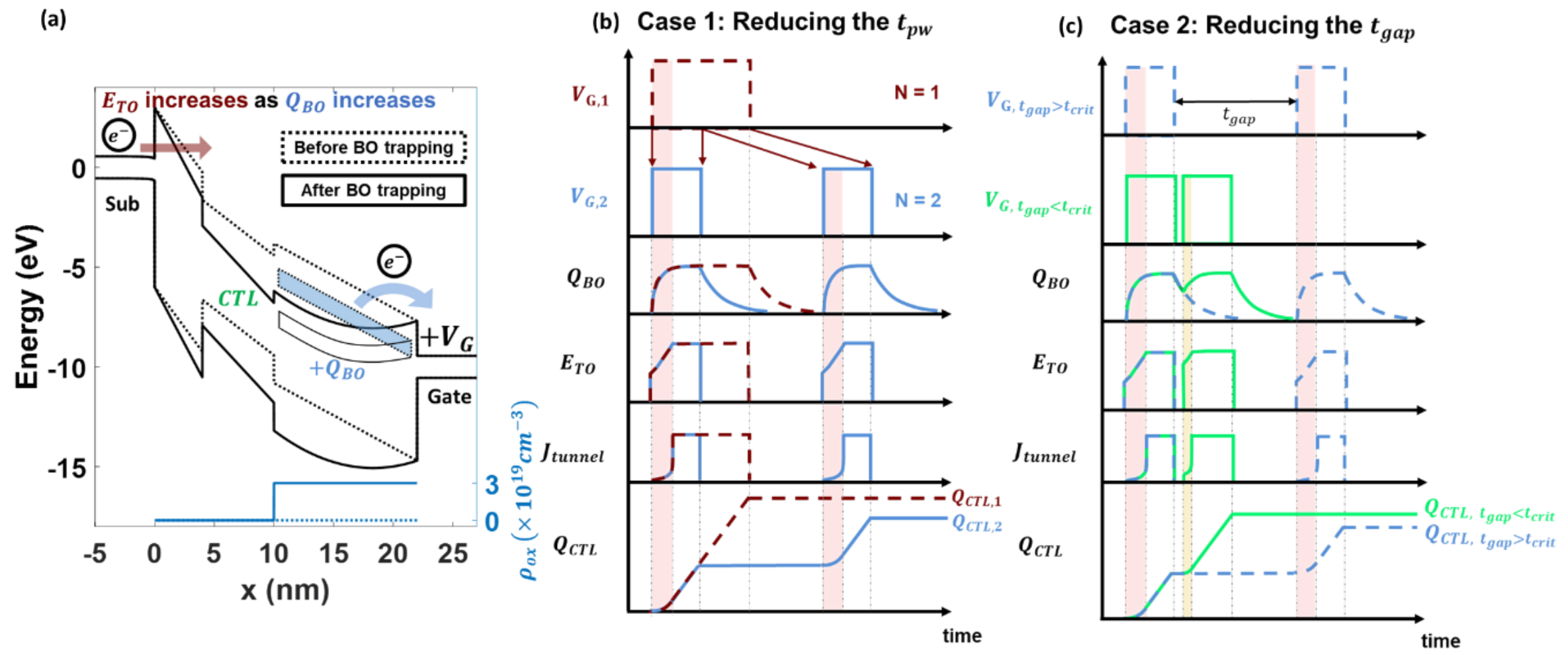

**Fig. 5. Model for $N$ and $t_{gap}$ dependence:** (a) Simulated band diagram during application of programming pulse before and after the BO is charged showing enhanced tunneling oxide field after BO trapping, (b) Qualitative effect of BO trapping time (shaded red) on accumulated charge in CTL when $t_{pw}$ is reduced ($N$ is increased), (c) Qualitative effect of reduced BO trapping time (shaded yellow) on accumulated charge in CTL when $t_{gap}$ is reduced.

Thus, the BO trapping dynamics during each programming pulse results in different charge stored in CTL and hence the measured $V_{T,N}$ vs. $N$ behaviors in Fig. 3. (a). Higher $N$ results in more dead-zones at the start of every pulse which ultimately shows up as the deficient accumulation of charge in CTL and hence the reduced $V_T$ shift. Similarly, when the programming pulse is removed:

1. The BO charges get de-trapped with a characteristic timescale $t_{crit}$.
2. There is a $t_{gap}$ before the next programming pulse arrives.
3. For $t_{gap} < t_{crit}$, there is residual BO charge before the next pulse arrives, this leads to a reduced dead-zone duration (shaded yellow in Fig. 5. (c)) as the BO trapping takes lower amount of time for the next pulse. Thus, the BO positive charge de-trapping dynamics explains the recovery of the charge accumulated in CTL as $t_{gap}$ reduces below $t_{crit}$ (Fig. 3. (b))
4. For $t_{gap} > t_{crit}$, the BO charge is fully de-trapped. For all larger $t_{gap}$, there is no interaction between pulses and measured $V_{T,N}$ saturates (Fig. 3. (b)).

It is expected that $t_{crit}$, i.e., the time it takes to remove all trapped BO charges reduces as $t_{pw}$ reduces (as observed in Fig. 3. (b)) since the amount of BO trapped charges per pulse may reduce as $t_{pw}$ reduces. It is important to note that the trapping/de-trapping dynamics is active when the programming pulses are applied and leave their signature in the accumulated charge in CTL. During the low voltage CV sweep for reading the stored charge in CTL and hence $V_T$, the BO traps are inactive due to the slow reading timescale ($\sim 1\ ms$) and low voltages during reading.

Now, we can appreciate why higher BO thicknesses show higher $V_{T,N}$ degradation with $N$ (Fig. 4. (c)). Thicker BO has more traps to fill which increases the role of trapping time and dead-zones compared to the actual tunneling time associated with higher $N$ or lower $t_{pw}$. Smaller pulse widths spend a smaller fraction of their on-duration at highly charged BO (i.e., when high non-volatile tunneling based storage is active) than longer pulse widths. This effect is more pronounced when there are more traps as is the case for thicker BO splits.

## 3.3. Estimating trapping/de-trapping timescales

We have established the mechanism for non-idealities in charge accumulation in CTF for time-fragmented pulse trains using the positive charge trap/de-trap dynamics in the BO. In this Section, we estimate the timescales of trapping and de-trapping of charges from BO using pulsing measurements.

### 3.3.1. De-trapping timescales

Charges in CTF, de-trap from the BO at zero gate bias, i.e., during the gap between consecutive pulses. Fig. 3. (b) shows that for large $t_{gap}$, the final $V_{T,N}$ at the end of pulse trains is saturated representing that all charges are de-trapped from the BO after every pulse for these $t_{gap}$. However, below a $t_{crit}$, the $V_{T,N}$ shows strong log-$t_{gap}$ dependence. This indicates the interaction of consecutive pulses through incomplete de-trapping or residual trapped charges in the BO after every pulse for $t_{gap} < t_{crit}$. Hence, the $t_{crit}$ defined and calculated in Section 3.1. is an estimate for the de-trapping timescale (Fig. 3. (c)) or $t_{de-trap} = t_{crit}$.

### 3.3.2. Trapping timescales

To estimate trapping timescales, we define a total pulse input writing time, $T_{tar}$, and an effective non-volatile writing time, $T_{NV}$, that achieve a particular charge storage in CTL, and thus a particular $V_{T,tar}$. If $n_{req}$ pulses of width $t_{pw}$ are needed to achieve $V_{T,tar}$, then:

$$T_{tar} = t_{pw}.n_{req} \tag{1}$$

Out of the total $T_{tar}$, a part of it lies in the dead-zones or the trapping duration, i.e., $t_{trap}$ per pulse. Hence it follows that the effective non-volatile writing time:

$$T_{NV} = T_{tar} - n_{req}t_{trap} = \left(t_{pw} - t_{trap}\right) \times n_{req} \tag{2}$$

Now, $t_{trap}$ can be calculated as:

$$t_{trap} = t_{pw} - T_{NV}/n_{req} \tag{3}$$

To measure how $V_T$ increases as consecutive pulses are applied, we use the same setup as in Section 2.2. (Fig. 2. (b)) but add a CV read at particular intermediate pulse numbers. This measures how the $V_{T,n}$ starts from fixed initial $V_{T,0}$ and approaches final $V_{T,N}$ with pulse number ($n$ is the incremental pulse number and $N$ is the total number of pulses). The measurements of this experiment are shown in Fig. 6. (a). Using these experiments, the methodology for extracting the trapping timescales is described as follows:

1. Choose a target $V_{T,tar}$ and calculate the number of pulses, $n_{req}$, required to reach that target $V_{T,tar}$ for different $t_{pw}$ (Fig. 6. (b) extracted from Fig. 6. (a)).
2. Find the intercept of the $n_{req}$ vs. $t_{pw}$ curve with a fixed $n_{req}$ (Fig. 6. (b)). We choose a moderately large value of $n_{req}$ for intercept (say $n_{req} = 200$, the intercept being $t_{pw,200}$) so that the pulse number quantization does not result in significant errors in $t_{trap}$ calculation.
3. Use the data of $V_{T,1}$, i.e., the one-shot $V_T$ after the first pulse of width $t_{pw}$ has been applied to extract $T_{NV}$ for the chosen $V_{T,tar}$ (dashed box in Fig. 6. (a), and Fig. 6. (c)). This method assumes that $T_{NV} = t_{pw} \gg t_{trap}$ and hence is expected to hold true only for higher $V_{T,tar}$ values typically.
4. Use the extracted $T_{NV}$ and $t_{pw,200}$ and equation (3) to calculate $t_{trap}$. This value is an average per pulse calculated for a target $V_{T,tar}$. Repeat the process for different $V_{T,tar}$ to obtain $t_{trap}$ as a function of $V_{T,tar}$.

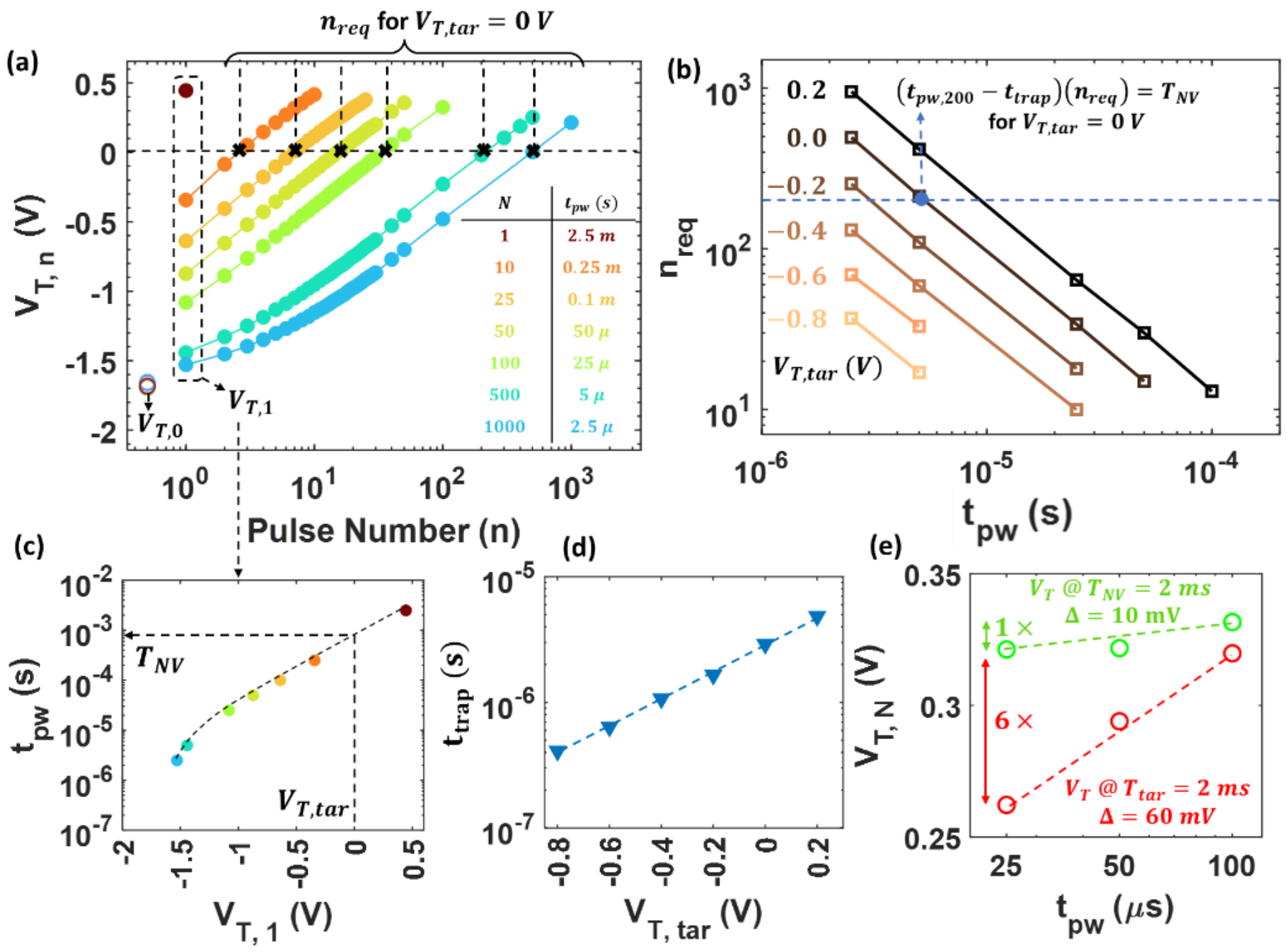

**Fig. 6. Estimating trapping timescales:** (a) $V_{T,n}$ (filled circles) as a function of pulse number ($n$) for pulse trains of different $t_{pw}$ (total train length, $N$) (open circles denote the initial $V_{T,0}$), (b) Required number of pulses, $n_{req}$, of a given $t_{pw}$ to reach a target $V_{T,tar}$, (c) $V_{T,1}$ after the application of first pulse of different $t_{pw}$, gives an estimate of $T_{NV}$, effective time for non-volatile charge storage in CTL (d) trapping timescales as a function of $V_{T,tar}$ estimated from $t_{pw,200}$ intercept of curves in (b) and $T_{NV}$ extracted from (c), (e) $V_{T,N}$ achieved for different $t_{pw}$ for $T_{tar} = 2\ ms$ (red) and for $T_{NV} = 2\ ms$ (green), shows $6\times$ improvement in charge accumulation deficiency.

The methodology is applied to $V_{T,tar} > -0.8\ V$ with $T_{NV} > 100\ \mu s$ (Fig. 6. (a)-(c)). The calculated $t_{trap}$ are shown in Fig. 6. (d). These are much smaller than the chosen $T_{NV}$ and hence the assumption in step 3 holds true. The calculated $t_{trap}$ (Fig. 6. (e)) are also much smaller than the $t_{de-trap}$ (Fig. 3. (c)) since trapping happens at high write voltages compared to de-trapping at zero gate bias. The $V_T$ achieved at the end of a $T_{tar} = 2\ ms$ is shown in Fig. 6. (e) for different $t_{pw}$. There is a $V_{T,N}$ reduction as expected for smaller $t_{pw}$ (higher $N$) despite the same writing time, $T_{tar}$. However, if we plot the $V_{T,N}$ achieved due to an effective writing time of $T_{NV} = 2\ ms$ for different $t_{pw}$, the charge accumulation deficiency is significantly eliminated ($6\times$ lower – green vs. red in Fig. 6. (e)). Thus, the reduced $V_T$ shift with $N$ observed in Section 3.1. can be primarily accounted for due to the blocking oxide trapping timescales.

## 3.4. Impact of CTF non-idealities

We have demonstrated that charge accumulation and hence $V_T$ shift is not conserved with $N$ (or $t_{pw}$), and $t_{gap}$. There are three key impacts when employing CTF as a memory array to implement RPU architecture (introduced in Section 1):

1. The stochastic encoding error can be reduced with the increase in $N$ where relative ***random*** error reduces as $\sigma/\mu = 1/\sqrt{N}$ (Fig. 7.) [7]. However, the non-idealities of dead period produce a maximum $N$ above which the $V_T$ shift becomes ineffective. Thus, there is a limited $N$ to improve error.
2. The $N$ dependence would add a ***systematic*** error in $V_T$ shift vs. pulse number curve. The fractional error in final $V_T$ shift vs. $N$ is shown in Fig. 7. assuming $N = 1$ as the ideal. However, the $t_{gap}$ is randomly changing for every bitstream. Hence, it adds stochastic noise to the final update. These systematic and random errors due to CTF need to be added to the existing random error due to stochastic pulse streams to produce a compounded (algorithm-device) non-ideality for the weight update.

3. In case different $N$ are used for simulated annealing like stochastic gradient descent [17], i.e., coarse updates at the beginning followed by finer weight updates, then it is necessary for the model of $V_T$ shift to include $N$ and $t_{gap}$. Such sophisticated models and analyzing their impact on training algorithms would be performed elsewhere in future work.

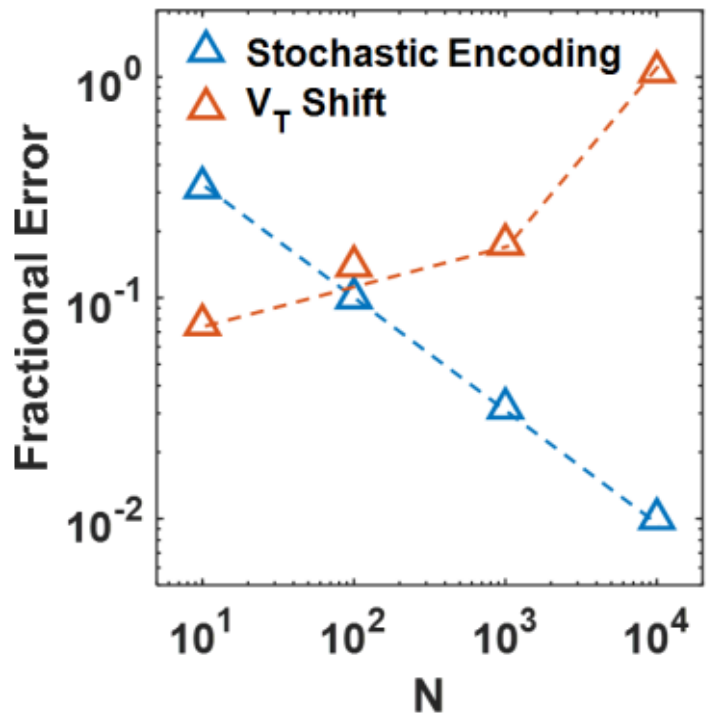

**Fig. 7. Fractional error:** Fractional error due to the stochastic encoding in pulse trains of length $N$ and $V_T$ shift error due to pulse division

# 4. Conclusion

In this paper, we measure the non-idealities in CTF charge accumulation which causes a reduced $V_T$ shift when the total ON-time, $T_{ON}$, is divided into $N$ discrete program pulses. This division is a common approach for many in-memory computing training architectures like the Resistive Processing Unit (RPU). It is observed that the $V_T$ shift reduces as $N$ increases and abruptly vanishes for yet smaller pulse widths. However, the $V_T$ shift gets recovered by reducing the $t_{gap}$ between consecutive pulses. A qualitative model to explain $N$ and $t_{gap}$ dependence based on blocking oxide trap dynamics is discussed and experimentally verified using process splits for the layers in the gate stack. A methodology to extract blocking oxide trapping/de-trapping timescales is discussed. A $6\times$ improvement in charge accumulation efficiency is demonstrated, provided the writing time is adjusted for the extracted trapping timescales. Finally, we present the context of Resistive Processing Unit (RPU) architecture for which the different errors need sophisticated modelling and presents the potential for controllable error stochastic gradient descent based on tuning $N$. The pulse division and inter-pulse interaction of CTF presented in this work can help the community design highly informed in-memory functional networks for hardware acceleration.

# Funding

This work is supported in parts by the Department of Science and Technology (DST) Nano Mission, Ministry of Electronics, and IT (MeitY) and Department of Electronics through the Nanoelectronics Network for Research and Applications (NNETRA) project, Govt. of India. Vivek Saraswat is supported by Prime Minister's Research Fellowship, Govt. of India.